\documentclass[letterpaper, 10 pt, conference]{ieeeconf}  % Comment this line out if you need a4paper
\IEEEoverridecommandlockouts                              % This command is only needed if 
\usepackage{times}
\usepackage{epsfig}
\usepackage{graphicx}
\usepackage{amsmath}
\usepackage{amssymb}

\usepackage{url}   % url link
\usepackage{multirow}
\usepackage{hhline}
\usepackage{amsmath}
\usepackage{adjustbox}
\usepackage{cite}

\usepackage{booktabs}
\usepackage{multirow}
\usepackage{graphicx}
\usepackage{subfig}
\usepackage{import}
\usepackage{siunitx}
\usepackage{pifont}% http://ctan.org/pkg/pifont
\newcommand{\cmark}{\ding{51}}%
\usepackage[dvipsnames]{xcolor}

\usepackage{import}

\usepackage{verbatim}

\newcommand{\figref}[1]{Fig\onedot~\ref{#1}}
\newcommand{\equref}[1]{Eq\onedot~\eqref{#1}}
\newcommand{\secref}[1]{Sec\onedot~\ref{#1}}
\newcommand{\tabref}[1]{Tab\onedot~\ref{#1}}

\newcommand{\thickhline}{%
    \noalign {\ifnum 0=`}\fi \hrule height 1pt
    \futurelet \reserved@a \@xhline
}

\newcommand{\onedot}{\futurelet\@let@token\@onedot}
\def\onedot{.}
\def\eg{\emph{e.g.~}}

\def\ie{\emph{i.e.~}}

\def\wrt{w.r.t\onedot~} 

\def\etal{\emph{et al.}}

\renewcommand{\baselinestretch}{0.94}

\usepackage[final]{hyperref}
\hypersetup{
  colorlinks   = true, %Colours links instead of ugly boxes
  urlcolor     = blue, %Colour for external hyperlinks
  linkcolor    = blue, %Colour of internal links
  citecolor   = red %Colour of citations
}
\begin{document}
\title{\LARGE \bf
Deep Keypoint-Based Camera Pose Estimation with Geometric Constraints
}

\author{
% \small
You-Yi Jau$^{*1}$ \\
% University of California, San Diego\\
{\tt\small yjau@eng.ucsd.edu}
\and
Rui Zhu$^{*1}$ \\
% UC San Diego \\
{\tt\small rzhu@eng.ucsd.edu}
\and
Hao Su$^{1}$ \\
% UC San Diego \\
{\tt\small haosu@eng.ucsd.edu}
\and
Manmohan Chandraker$^{1}$ \\
% UC San Diego \\
{\tt\small mkchandraker@eng.ucsd.edu}
% {pin: 152045}
\thanks{*Equal contribution}% <-this % stops a space
\thanks{$^{1}$ The authors are with University of California, San Diego}% <-this % stops a space
% \thanks{1.Affiliated in University of California, San Diego}% <-this % stops a space
}
\maketitle

%\thispagestyle{empty}
%%%%%%%%% ABSTRACT
\begin{abstract}
    % \ishit {hello ishit}
    Estimating relative camera poses from consecutive frames is a fundamental problem in visual odometry (VO) and simultaneous localization and mapping (SLAM), where classic methods consisting of hand-crafted features and sampling-based outlier rejection have been a dominant choice for over a decade. Although multiple works propose to replace these modules with learning-based counterparts, most have not yet been as accurate, robust and generalizable as conventional methods. In this paper, we design an end-to-end trainable framework consisting of learnable modules for detection, feature extraction, matching and outlier rejection, while directly optimizing for the geometric pose objective. We show both quantitatively and qualitatively that pose estimation performance may be achieved on par with the classic pipeline. Moreover, we are able to show by end-to-end training, the key components of the pipeline could be significantly improved, which leads to better generalizability to unseen datasets compared to existing learning-based methods.\\
\end{abstract}

%%%%%%%%% BODY TEXT

\section{Introduction}
% ```
% - SLAM is important
%   - SLAM rgbd -> rgb
% - Traditional SLAM (Orb-slam) pipeline

% - Camera motion estimation is important
% - multiview geometry
% - Superpoint and Deep fundamental matrix
% - In our work, we ...
% - with the following contributions:
% (Simultaneously localization and mapping (SLAM) has been explored for decades. )
% ```

Camera pose estimation has been the key to Simultaneous Localization and Mapping (SLAM) systems. To this end, multiple methods have been designed to estimate camera poses from input image sequences, or in a simplified setting, to get the relative camera pose from two consecutive frames. Traditionally a robust keypoint detector and feature extractor, \eg SIFT \cite{lowe_distinctive_2004}, coupled with an outlier rejection framework, \eg RANSAC \cite{fischler_random_1981}, has dominated the design of camera pose estimation pipeline for decades.

\begin{figure*}
    \centering
    \includegraphics[page=2, width=500pt, height=138pt, keepaspectratio]{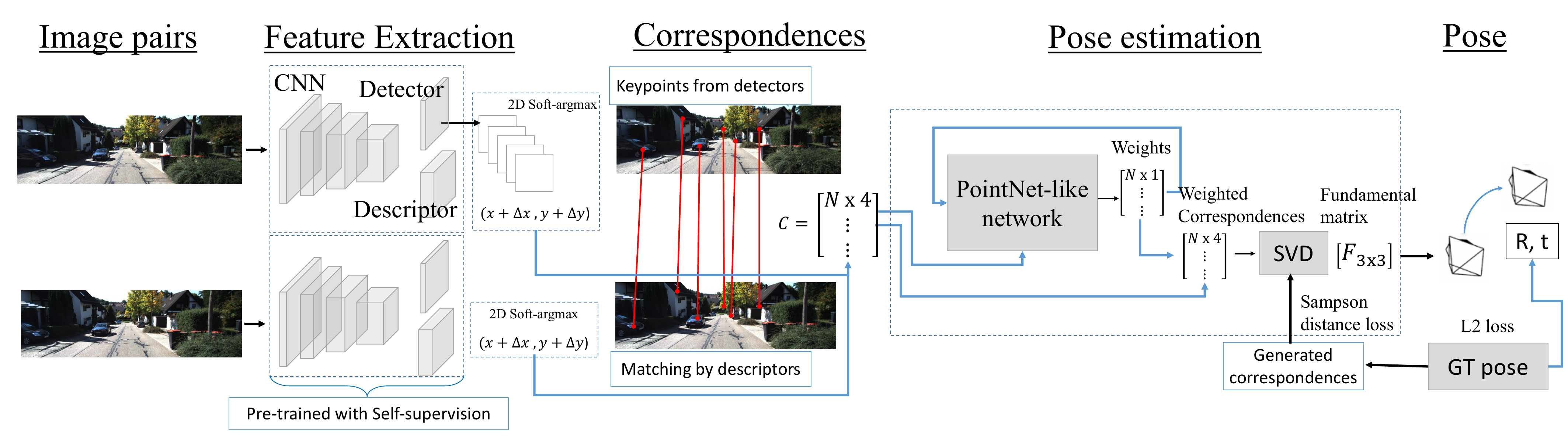}\label{fig:pip_over}  % ppt: 72cm x 20cm
    % \includegraphics[page=3, width=240pt, height=150pt]{latex/pdf/pipeline_overview_v2.pdf}\label{fig:pip_over}
    % \subfloat[Sequence 09.]{}
    \caption{\textbf{Overview of the system.} A pair of images is fed into the pipeline to predict the relative camera pose. Feature extraction predicts detection heatmaps and descriptors for finding sparse correspondences. Local 2D \texttt{Softargmax} is used as a bridge to get subpixel prediction with gradients. Matrix $\mathbf{C}$ of size $N \times 4$ is formed from correspondences. 
    $\mathbf{C}$ is the input for pose estimation, where the PointNet-like network predicts weights for all correspondences. Weighted correspondences are passed through SVD to find fundamental matrix $F$, which is further decomposed into poses. Ground truth poses (GT poses) are used to compute L2 loss between rotation and translation (\textbf{pose-loss}). Correspondences generated from GT poses are used to compute fundamental matrix loss (\textbf{F-loss}). See more details in \secref{Method}.}
    \label{fig:pip_supervision}
\end{figure*}

Recently there have been efforts to bring deep networks to each step of the pipeline, specifically keypoint detection~\cite{detone_superpoint:_2017, ono_lf-net:_2018, yi_lift:_2016}, feature extraction \cite{detone_superpoint:_2017,ono_lf-net:_2018} and matching~\cite{brachmann2017dsac,ono_lf-net:_2018,detone_superpoint:_2017}, as well as outlier rejection \cite{brachmann2017dsac, dsac++, deepf}. The potential benefit is being able to handle challenges such as textureless regions by incorporating data-driven priors. However, when combining such components to replace the classic counterparts, conventional SIFT-based camera pose estimation still significantly outperforms them by a considerable margin. This could be attributed to three basic challenges for learning-based systems. First, these learning-based methods have been individually developed for their own purposes, but never been trained and optimized end-to-end for the ultimate purpose of getting better camera poses. Geometric constraints and the final pose estimation objective are not sufficiently incorporated in the pipeline. % Instead, these methods optimize for mostly better fundamental matrices, which do not necessarily lead to better camera pose. 
Second, learning-based methods have over-fitting nature to the domains they are trained on. When the model is applied to a different dataset, the performance is often inconsistent across various datasets compared to SIFT and RANSAC methods. Third, our evaluation shows that existing learning-based feature detectors, which serve at the very beginning of the entire pipeline, are significantly weaker than the hand-crafted feature detectors (\eg SIFT detector). This is because obtaining training samples with accurate keypoints and  correspondences, at the level to surpass or just match the subpixel accuracy of SIFT, is tremendously difficult.

In face of these issues when naively putting existing learning-based methods together, we propose the end-to-end trained framework for relative camera pose estimation between two consecutive frames (\figref{fig:pip_supervision}). Our framework integrates learnable modules for keypoint detection,  description and outlier rejection inspired by the geometry-based classic pipeline. The whole framework is trained in an end-to-end fashion with supervision from ground truth camera pose, which is the ultimate goal for pose estimation.
% in the hope for better robustness and generalizability. The whole framework is trained in an end-to-end fashion with supervision from ground truth camera pose, meanwhile incorporating geometry-based pose objective, so as to encourage feature detection, feature extraction and matching to be better coupled for the eventual camera pose estimation task. 
Particularly, in facing the third challenge of requiring accurate keypoints for feature detector training, we introduce a \texttt{Softargmax} detector head in the pipeline, so that the final pose estimation error could be back-propagated to provide subpixel level supervision. 

Experiments show that the end-to-end learning can drastically improve the performance of existing learning-based feature detectors, as well as the entire pose estimation system. We show that our method outperforms existing learning-based pipelines by a large margin, and performs on par with the state-of-the-art SIFT-based methods. We also demonstrate the significant benefit of generalizability to unseen datasets compared to learning-based baseline methods. We evaluate our model on KITTI \cite{kitti} and ApolloScape \cite{huang_apolloscape_2019} datasets and demonstrate not only quantitatively but also qualitatively. That is, by training end-to-end, we are able to obtain relatively balanced keypoint distribution corresponding to appearance and motion patterns in the image pair.

To summarize, our contributions include:
\begin{itemize}
    \item We propose the keypoint-based camera pose estimation pipeline, which is end-to-end trainable with better robustness and generalizability than the learning-based baselines.
    \item The pipeline is connected with the novel \texttt{Softargmax} bridge, and optimized with geometry-based objectives obtained from correspondences.
    \item The thorough study on cross-dataset setting is done to evaluate generalization ability, which is critical but not much discussed in the existing works.
\end{itemize}
% Finally, we also demonstrate that end-to-end training benefits all stages of our pipeline, thereby also leading to better keypoint detection and matching.

We describe our pipeline in detail in \secref{Method} with the design of the loss functions and training process. We show the quantitative results of pose estimation and qualitative results in \secref{Experiments}. Code will be made available at \url{https://github.com/eric-yyjau/pytorch-deepFEPE}.
% \url{https://github.com/eric-yyjau/deepFEPE_pytorch.git}.
% Ablation study is shown in \secref{Ablation}.

% Fig. ? shows the examples of input and output. Fig. \ref{fig:pip1} shows the overview of our system.

% . Keypoints are extracted from the heatmap using non-maximum suppression. Patches around keypoints are extracted and sent through soft-argmax to predict subpixel residual. After matches are discovered by matching the 2 sets of descriptors around keypoints, $N$ correspondences are formed.
    % \rui{give it a name}
%   compute sampson's distance with estimated fundamental matrix. The detailed network structure is shown in Figure.
\section{Related Work}

% \textbf{Geometry based SLAM}
\noindent \textbf{Geometry-based visual odometry}
Visual odometry (VO) is a well-established field \cite{engel_semi-dense_2013,newcombe_dtam:_2011,forster_svo:_2014}, which estimates camera motion between image frames. 
This line of research can be separated into two main groups, feature-based methods and direct methods. For feature-based methods, \eg \cite{Geiger2011IV, mur-artal_orb-slam:_2015}, sparse keypoints for images are detected and described in order to form a set of correspondences. The correspondences are then used for pose estimation using 8-point algorithm \cite{hartley_defense_1997}, PnP \cite{lepetit_epnp:_2009}, or optimized jointly with pose using bundle adjustment \cite{triggs_bundle_2000}. Due to the presence of localization noise and outliers, RANSAC\cite{fischler_random_1981} is a popular choice for outlier rejection. However, the method struggles in case of textureless or repetitive patterns, where keypoints are noisy and difficult to match, as it only uses sparse features across the image and strives to find a good subset of them. This issue motivated the direct methods\cite{engel_lsd-slam:_2014,newcombe_dtam:_2011}, which maximizes photometric consistency over all pixel pairs. However, the method suffers in dynamic scenes or challenging lighting environments. Methods combining sparse feature-based methods and direct methods are proposed in recent years, \eg\cite{forster_svo:_2014, engel_semi-dense_2013, engel_direct_2016}, and in addition, loop closure and bundle adjustment (BA) have been applied to extend VO to simultaneous localization and mapping (SLAM).%, where ORB-SLAM\cite{mur-artal_orb-slam:_2015} is considered as a classic baseline.
% The pipeline consists of multiple modules with heuristic or hand-crafted parameters, such as the thresholds for RANSAC algorithm.
% ORB-SLAM, LSD-SLAM
%  In direct method,  It can work better in textureless situations, but is more unstable to dynamic objects due to the lack of outlier rejection. 
% To solve the shortness of both methods, learning-based system or modules of SLAM has some progress (CNN-SLAM, bundle adjustment), but a robust system has not yet to be seen.\\

\noindent \textbf{Learning-based visual odometry}
Deep learning for VO has developed rapidly in recent years, \eg\cite{tateno_cnn-slam:_2017, wang_deepvo:_2017, li_undeepvo:_2018, kendall_posenet:_2016, tang_ba-net:_2018, zhou2018deeptam, li2019pose, NIPS2019_scsfmlearner, ferrari_relocnet_2018}, taking advantage of convolutional neural networks (CNN) for better adaptation to specific domains. For the monocular camera setting, CNN-SLAM \cite{tateno_cnn-slam:_2017} claims that learned depth prediction helps in textureless regions and corrects the scale. PoseNet\cite{kendall_posenet:_2016} utilizes CNN to predict a 6 DoF global pose and claims the ability to adapt to a new sequence with fine-tuning. To take advantage of temporal information, DeepVO \cite{wang_deepvo:_2017, wang2018end} proposes to use recurrent neural networks (RNN) to predict poses along the sequence. The most recent work is \cite{xue2019beyond} which brings learnable memory and refinement modules into the framework. For the sparse feature-based category, some work has been done using learning-based methods, \eg\cite{kang_df-slam:_2019, detone_superpoint:_2017,detone_self-improving_2018,zhan_vorevisited_2019}. In \cite{kang_df-slam:_2019}, the feature descriptor of a two-layer shallow network is combined with the SLAM pipeline. In addition, SuperPoint \cite{detone_superpoint:_2017,detone_self-improving_2018}, which is a learned feature extractor, is combined with BA to update the stability score for each point. However, learned feature extractor is employed in an off-the-shelf manner, which may not be optimal from an end-to-end perspective. 

% To solve the shortness of both methods, learning-based system or modules of SLAM has some progress (CNN-SLAM, bundle adjustment), but a robust system has not yet to be seen.

% using 2D-2D correspondences (8-point algorithm), 2D-3D correspondences (PnP), or optimized jointly pose and point locations using 3D-3D correspondences (bundle adjustment)
\noindent \textbf{Learning-based feature extraction and matching}
Feature extraction, which consists of keypoint detection and description, has been utilized in a variety of vision problems. 
Traditionally, detectors \cite{trajkovic_fast_1998,rublee_orb:_2011,lowe_distinctive_2004} and descriptors \cite{lowe_distinctive_2004,rublee_orb:_2011} are mostly designed by heuristics. SIFT \cite{lowe_distinctive_2004}, which utilizes the Gaussian feature pyramid and descriptor histograms, has achieved success over the past decade. 
In recent years, deep learning has been utilized to build up feature extractors, \eg\cite{yi_lift:_2016,ono_lf-net:_2018,detone_superpoint:_2017}.
% Superpoint, LF-net, LIFT
To our knowledge, LIFT\cite{yi_lift:_2016} is the first end-to-end pipeline, which consists of a SIFT-like procedure with sliding window detection and is trained on ground truth generated from SIFT and SfM\cite{wu_towards_2013}. LF-net\cite{ono_lf-net:_2018} optimizes keypoints and correspondences with gradients using ground truth camera poses and depth. SuperPoint\cite{detone_superpoint:_2017} proposes a self-supervised pipeline to train detectors and descriptors at the same time and beats SIFT in HPatches\cite{balntas_hpatches:_2017} evaluation, with some follow-up works \cite{tang_self-supervised-3d_2019, christiansen_unsuperpoint_2019}. However, all of the feature extractors are not optimized in together with the overall VO system, leading to suboptimal performance. Also, their evaluation metrics, \ie matching score, does not necessarily reflect the performance of pose estimation in a VO task.

\begin{figure*}
    \centering
    \includegraphics[page=2, width=480pt, height=137pt]{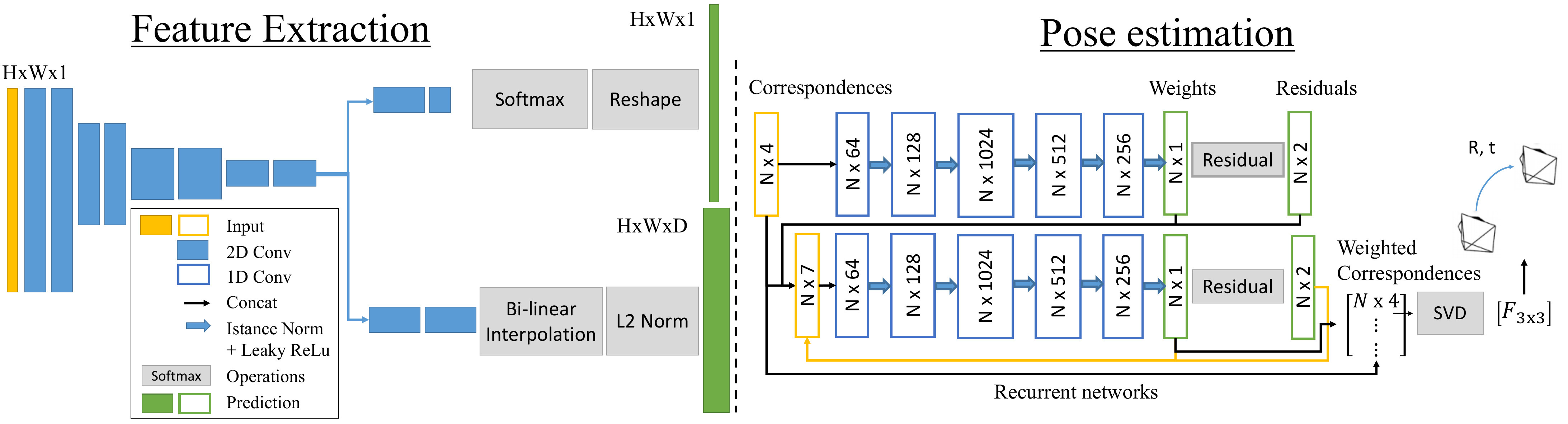}
    \caption{ \textbf{ Network structure of our feature extraction (FE) and pose estimation(PE) modules.}
    % \yy{add network description}
    FE module\cite{detone_superpoint:_2017} has VGG-like structure, with gray images as input, and detection and description heatmaps as output. PE module has $N$ correspondences ($\mathbf{C}$) as input, with several layers of 1D convolution to initialize weights and compute residuals. The weights, residuals and correspondences are fed into the RNN with the same structure for D iterations (D=5). From the final correspondences with weights, the fundamental matrix is estimated.
    % passing through 8 layers of CNN encoder with kernel size 3, with depth 64, 64, 128, 128, 128, 128, 256, 256 respectively. The output of the encoder is fed into decoder head depth 65 and one 
    }
    \label{fig:pip_network}
\end{figure*}

\noindent \textbf{Learning-based camera pose estimation}
% Berkeley paper, DSAC, learning from correspondences
% This method suffers from the bias of training data and dynamic objects.
Learning-based methods for camera pose estimation have been gaining attention in recent years. Following direct methods, \cite{zhou_unsupervised_2017,yin_geonet:_2018, wang2018learning, godard2019digging,ranjan2019competitive, li_undeepvo:_2018, yin2018geonet} take advantage of geometric constraints of 3D structure, and jointly estimate depth and pose in an unsupervised manner using photometric consistency. Poursaeed \etal \cite{poursaeed_deep_2018} uses Siamese networks \cite{koch2015siamese} to regress the fundamental matrix between left and right views through the sequence. For feature-based pipeline, \cite{brachmann2017dsac} makes a differentiable sampling-based version of RANSAC, while \cite{deepf,moo_yi_learning_2018} utilize PointNet-like architecture\cite{qi_pointnet:_2016} to weigh each input correspondence and subsequently solve for the camera pose. These methods retain the mathematical and geometric constraints from classic methods, therefore can be more generalizable than direct prediction from image appearance. 
% This setting in replacement of RANSAC, not only proves to have robust outlier rejection, but can enable gradient flowing back to the other modules as well. This building block make end-to-end training possible.

% The works done above show that end-to-end feature based VO can potentially benefit from geometric constraints enforced explicitly in the pipeline, which however have not been proved before. 

% The potential motivates us to investigate in the line of research. By connecting state-of-the-art learning based feature extraction and matching with relative pose estimation, the pipeline optimized end-to-end can reach a new baseline for VO. It can be further extended into robust SLAM pipeline by combining the strength of networks and theoretical constraints.
% \import{sections/}{notation.tex}

\section{Method} \label{Method}

\subsection{Overview}
We propose a deep feature-based camera pose estimation pipeline called \texttt{DeepFEPE} (Deep learning-based Feature Extraction and Pose Estimation), which takes two frames as input and estimates the relative camera pose. 
The pipeline mainly consists of two learning-based modules, for feature extraction and pose estimation respectively, as shown in \figref{fig:pip_supervision}. 

Instead of naive concatenation of the modules, careful designs are made for training \texttt{DeepFEPE} end-to-end, which includes \texttt{Softargmax} \cite{chapelle_gradient_2010} detector head and geometry-embedded loss function. The \texttt{Softargmax} detector head equips the feature detection with sub-pixel accuracy, and enables gradients from pose estimation to flow back through the point coordinates. For loss function, we not only regress a fundamental matrix, but also directly constrain the decomposed poses by enforcing a geometry inspired L2 loss on the estimated rotation and translation, which leads to better prediction and generalization ability as shown in \secref{Experiments}. We include more details for \texttt{DeepFEPE} and the network structures in \figref{fig:pip_network}.

\noindent \textbf{Notation} \label{notation}
We refer to the pair of images as $I, I' \in \mathbb{R}^{H\times W}$, the transformation matrix from frame $i$ to $j$ as $T_{ij} = [R| t]$, where $R \in \mathbb{R}^{3 \times 3}$ is the rotation matrix and $t \in \mathbb{R}^{3 \times 1}$ is the translation vector.
We refer to a point in 2D image coordinates as $p \in \mathbb{R}^2$, where $p = [u, v]$. 

\subsection{Feature Extraction (FE)} \label{feature_ext}
We use learning-based feature extraction (FE), namely SuperPoint \cite{detone_superpoint:_2017}, in our pipeline. SuperPoint is chosen as our base component because it is trained with self-supervision and demonstrated top performance for homography estimation in HPatches dataset \cite{balntas_hpatches:_2017}.
Similar to traditional feature extractors, \eg SIFT, SuperPoint serves as both the detector and descriptor, with the input gray image $I \in \mathbb{R}^{H\times W \times 1}$, and output keypoint heatmap $H_{det} \in \mathbb{R}^{H\times W \times 1}$ and descriptor $H_{desc} \in \mathbb{R}^{H\times W \times D}$.  SuperPoint consists of a fully-convolutional neural network with a shared encoder and two decoder heads as the detector and descriptor respectively, as shown in \figref{fig:pip_network}.
% We made 2 major adjustments for the original design. First, we apply softargmax at the output points to enable end-to-end training. Secondly, we use sparse loss to train the descriptor.
\subsubsection{\texttt{Softargmax} Detector Head}
To overcome the challenge of training end-to-end, we propose detector head with 2D \texttt{Softargmax}. In the original Superpoint, non-maximum suppression (NMS) is applied to the output of keypoint decoder $H_{det}$ to get sparse keypoints. However, the direct output from NMS only has pixel-wise accuracy and is non-differentiable. Inspired by \cite{ono_lf-net:_2018}, we apply \texttt{Softargmax} on the $5\times 5$ patches extracted from the neighbors of each keypoint after NMS. The final coordinate of each keypoint can be expressed as
% \[ 
% \begin{bmatrix} x', y'
% \end{bmatrix}
% = 
% \begin{bmatrix} x_0, y_0
% \end{bmatrix}
% +
% \begin{bmatrix} \delta x, \delta y
% \end{bmatrix}\]

\begin{equation}
(u', v') = (u_0, v_0) + (\delta u, \delta v),
\end{equation}
where in a given 2D patch,
% \[
\begin{equation}
 \delta u = \frac{\sum_{j}\sum_{i}e^{f(u_i,v_j)} i }{\sum_{j}\sum_{i}e^{f(u_i,v_j)}}, 
 \delta v = \frac{\sum_{j}\sum_{i}e^{f(u_i,v_j)} j }{\sum_{j}\sum_{i}e^{f(u_i,v_j)}}.
\end{equation}
% \delta v = \frac{\sum_{j}\sum_{i}e^{f(u_i,v_j)}*j }{\sum_{j}\sum_{i}e^{f(u_i,v_j)}}.
%  \delta x = \frac{\sum_{i,j}e^{f(x_i,y_j)}*i }{\sum_{i}e^{f(x_i,y_j)}} 
% \]
$f(u,v)$ denotes the pixel value of the heatmap at position $(u,v)$, and $i,j$ denotes the relative directions in x, y-axis with respect to the center pixel $(u_0,v_0)$. The integer-level keypoint $(u_0,v_0)$ is therefore updated to $(u', v')$ with subpixel accuracy.

The output of the \texttt{Softargmax} enables flow of gradients from the latter module to the front, in order to refine the coordinates for subpixel accuracy. To pre-train the FE module with \texttt{Softargmax}, we convolve the ground truth 2D detection heatmap with a Gaussian kernel $\sigma_{fe}$. The label of each keypoint is represented as a discrete Gaussian distribution on a 2D image. 

\subsubsection{Descriptor Sparse Loss}
To pre-train an efficient FE, we adopt sparse descriptor loss instead of dense loss. Original dense loss \cite{detone_superpoint:_2017} collects loss from all possible correspondences between two sets of descriptors in low resolution output, which creates a total of $(H_c \times W_c)^2$ of positive and negative pairs. Instead, we sparsely sample $N$ positive pairs, and $M$ negative pairs collected from each positive pair, forming $M \times N$ pairs of sampled correspondences. The loss function is the mean contrastive loss as described in \cite{detone_superpoint:_2017}.

% The loss function can be represented as 
% \begin{equation}
% L_{desc}(D, D') = L_{pos}(D, D') + L_{neg}(D, D')   
% \end{equation}
% where $H_{desc}, H'_{desc}$ are the dense descriptors and warped dense descriptors.
% % L_{desc}(D, D') = L_{matching}(D, D') + L_{non-matching}(D, D')

% \begin{equation}
% L_{pos}(D, D') = \frac{1}{N_{pos}} \sum | M_p - D(u_i, v_i) \cdot D(u_i', v_i') | 
% \end{equation}

% \begin{equation}
% L_{neg}(D, D') = \frac{1}{N_{pos}}\frac{1}{N_{neg}} \sum | D(u_i, v_i) \cdot D(u_i', v_i') - M_n |    
% \end{equation}\\

% where $N_{pos}$ denotes positive matching pairs and $N_{neg}$ denotes negative pairs for each positive pair. We use cosine distance for the pair of descriptors. 
% $M_p$ and $M_n$ are hyper-parameters for the margins.

% The overall loss function can be represented as
% \begin{equation}
% \begin{aligned}
% L(X,X', Y,Y', D, D') = \\
% L_{det}(X,Y) + L_{det}(X', Y') + \lambda L_{desc}(D, D')
% \end{aligned}
% \end{equation}

% We choose $M_p = 1$, $M_n = 0.2$, $N_{pos} = 600$, $N_{neg} = 100$, $\lambda = 1$.

% The output of the descriptor head follows original design, which is a dense high-dimensional heatmap. The descriptor of each keypoint is gathered by bi-linear sampling from the heatmap.

\subsubsection{Output of Feature Extractor}
We obtain correspondences for pose estimation from the sparse keypoints and their descriptors. To get the keypoints, we apply non-maximum suppression (NMS) and a threshold on the heatmap to filter out redundant candidates. The descriptors are sampled from $H_{desc}$ using bi-linear interpolation. With two sets of keypoints and descriptors, 2-way nearest neighbor matching is applied to form $N$ correspondences, forming an $N \times 4$ matrix as input for pose estimation.
% ratio test in SIFT, we use NN search (and filter out weak matchings with a threshold). 
% softargmax function
% \subsection{Feature matching}

\begin{table*}
    \scriptsize
    \centering
    \begin{tabular}{lllllllllll}
        \toprule
        \multicolumn{2}{c}{\multirow{2}{*} {Model References}} & \multicolumn{2}{c}{Feature extraction} & \multicolumn{2}{c}{Pose estimation} & \multicolumn{2}{c}{Loss} & \multicolumn{1}{c}{Training} \\
        \cmidrule(r){3-4} \cmidrule(r){5-6} \cmidrule(r){7-8} \cmidrule(r){9-9} \\
         Categories & Symbols & {Sift (Si)} & Superpoint (Sp) & {RANSAC (Ran)} & DeepF (Df) & {F-loss (f)} & Pose-loss (p) & End-to-end (end) \\
        \midrule
        \textbf{SIFT + RANSAC (Si-base)}              & Si-Ran       & \cmark                         &                 & \cmark                                 &            &                                  &               &                  \\  \midrule % \cline{2-2}
        \textbf{Superpoint + RANSAC (Sp-base)}        & Sp-Ran       &                                & \cmark               & \cmark                                 &            &                                  &               &                  \\ \midrule
        % \multirow{3}{*}
        \textbf{Baseline with Sift + DeepF }            & Si-Df-f      & \cmark                              &                 &            & \cmark          & \cmark                                &               &                  \\ % \cline{2-2}
        \textbf{(Si-models)}                            & Si-Df-p      & \cmark                              &                 &            & \cmark          &                                  & \cmark             &                  \\ % \cline{2-2}
                                                        & Si-Df-fp     & \cmark                              &                 &            & \cmark          & \cmark                                & \cmark             &                  \\ \midrule
        % \multirow{2}{*}
        \textbf{Ours - no end-to-end training }   & Sp-Df-f      &                                & \cmark               &                                   & \cmark          & \cmark                                &               &                  \\ % \cline{2-2}
        \textbf{(Sp-models)}                                              & Sp-Df-p      &                                & \cmark               &                                   & \cmark          &                                  & \cmark             &                  \\ \midrule
        % \multirow{3}{*}
        \textbf{Ours - with end-to-end training } & Sp-Df-f-end  &                                & \cmark               &                                   & \cmark          & \cmark                                &               & \cmark                \\ % \cline{2-2}
        \texttt{(DeepFEPE)}                                        & Sp-Df-p-end  &                                & \cmark               &                                   & \cmark          &                                  & \cmark             & \cmark                \\ % \cline{2-2}
                                                              & Sp-Df-fp-end &                                & \cmark               &                                   & \cmark          & \cmark                                & \cmark             & \cmark                \\ 
        \bottomrule
    \end{tabular}
\caption{\label{tab:exp_ablation_ref_table} \textbf{The reference table for modules and losses trained for experiments.} The table lists all the baselines used in \secref{sec:pose_est}. Baselines and our models are referred to by symbols, as they consist of different FE or PE modules trained using different losses.}
\end{table*}
% \subsection{Deep fundamental matrix} \label{DeepF}
\subsection{Pose Estimation (PE)} \label{DeepF}
%%% why choose this
% However, the main challenge is that RANSAC is not differentailable due to the step function for hard selection. 
Pose estimation takes correspondences as input to solve for the fundamental matrix. Instead of using a fully connected layer to regress fundamental matrix or pose directly as in \cite{zhou_unsupervised_2017,poursaeed_deep_2018, ferrari_relocnet_2018}, we embed geometric constraints, \ie sparse correspondences, into camera pose estimation. To create a  differentiable pipeline in replacement of RANSAC for pose estimation from noisy correspondences, we build upon the Deep Fundamental Matrix Estimation (DeepF) \cite{deepf}, and propose a geometry-based loss to train \texttt{DeepFEPE}.

\subsubsection{Existing Objective for Learning Fundamental Matrix}
DeepF \cite{deepf} formulates fundamental matrix estimation as a weighted least squares problem. The weights on the correspondences indicate the confidence of matching pairs, and are predicted using a neural network model with the PointNet-like structure. Then, weights and points are applied to solve for the fundamental matrix. 
Residuals of the prediction, as defined in \cite{deepf}, are obtained from the mean Sampson distance \cite{sampson} of the input correspondences.
% the fundamental matrix from correspondences sampled 
% from the ground truth fundmental matrix. 
The correspondences, weights, residuals are fed into the model recurrently to refine the weights. 
To be more specific, the residuals $r(\mathbf{p_i}, \mathbf{F})$ are calculated as following:

% \[g(x) = (x^T, ) \]   
\begin{equation}
r(\mathbf{p_i}, \mathbf{F}) = 
| \mathbf{\hat{p}_i}^{T}\mathbf{F}\mathbf{\hat{p}_i}'|
(\frac{1}{\|\mathbf{F}^{T}\mathbf{\hat{p}_i} \|_2}+\frac{1}{\|\mathbf{F}\mathbf{\hat{p}_i}' \|}_2),
\label{eq:sampson}
\end{equation}

\noindent where $\mathbf{p} = (u, v, 1)$ and $ \mathbf{p'} = (u', v', 1)$ denote a pair of correspondences in the homogeneous coordinates.
% The weights are updated using an estimator with point features and residuals as input. 

Following \cite{deepf}, the loss is defined as epipolar distances from \textit{virtual} points on a grid, to their corresponding epipolar lines, which are generated from ground truth fundamental matrix. It is abbreviated as \textbf{F-loss} in the following sections.

% \[r = (13)\]
% \begin{equation}
%   L = \frac{1}{N_{gt}}\sum \sum min(r(p_i^{gt}, g(x^j)), \gamma)
% \end{equation}

\subsubsection{Geometry-based Pose Loss}
Due to the fact that a good estimation in epipolar space does not guarantee better pose estimation, we propose a geometry-based loss function by enforcing a loss between estimated poses and ground truth poses. 
% The reason is that the vector space of the fundamental matrix is larger than that of the pose. Correct fundamental matrix doesn't lead to correct rotation and translation. 
The estimated fundamental matrix is converted into the essential matrix using the calibration matrix and further decomposed into 2 sets of rotation and 2 translation matrices.
% , with only a correct one that make the points in the front of both cameras.
By picking the one camera pose where all points are in front of both cameras (which gives lowest error among all 4 possible combinations of poses), we obtain the rotation in quaternions \cite{zhang_quaternions_1997} and translation vectors, and compute L2 loss as our geometry-based loss.
% The input correspondences are used to check the validity of the pose. 
% Then, loss terms are the minimum value between different sets of poses
Then, loss terms $\mathcal{L}_{rot}$ and $\mathcal{L}_{trans}$ are collected from the pose with minimum L2 loss.
\begin{equation}
%   \mathcal{L}_{rot} = min(\|q(\mathbf{R}_{est\_i}^T \mathbf{R}_{gt})\|_2), i=[1,2],
    \mathcal{L}_{rot} = min(\|q(\mathbf{R}_{est\_i}) - q(\mathbf{R}_{gt})\|_2), i=[1,2],
\end{equation}
\begin{equation}
   \mathcal{L}_{trans} = min(\|\mathbf{t}_{est\_i} - \mathbf{t}_{gt}\|_2), i=[1,2],
\end{equation}
where $R, t$ are decomposed from the essential matrix, and $q(.)$ converts the rotation matrix into a quaternion vector. 

\noindent The final loss is followed,
% \[L2 loss of pose in quarternion\]
\begin{equation}
\begin{aligned}
\mathcal{L}_{pose} = &min(\mathcal{L}_{rot}(\mathbf{R}_{est}, \mathbf{R}_{gt}), c_{r}) + \\
            &\lambda_{rt} * min(\mathcal{L}_{trans}(\mathbf{t}_{est}, \mathbf{t}_{gt}), c_{t}),
\end{aligned}
\end{equation}

% \begin{equation}
%   [R_{est}|t_{est}]_i = decompose(E_{est}), i = [1,2] 
% \end{equation}
% \begin{equation}
%     E_{est} = K^{T}FK    
% \end{equation}
\noindent where $c_r$ and $c_t$ are clamping constants for losses to prevent gradient explosion. 
The geometry-based loss is abbreviated as \textbf{pose-loss} in the following sections.
% We have multiple options when decomposing essential matrix $E_{est}$ into $[R_{est}|t_{est}]$. Instead of checking the projection of points, we naively choose the one with minimum l2 loss.

% \subsection{Loss functions}
% The final loss function is the loss from pose, in the form of quarternion. The loss is used to train both pose estimation model and feature extractor to get the best result.

\subsection{Training Process} \label{training_process}
After initializing both FE and PE modules respectively, we train the pipeline end-to-end.
%  based on the correspondences from FE
Our FE module is trained using a self-supervised method \cite{detone_superpoint:_2017}. The keypoint detector is initialized by synthetic data, which can be used to generate pseudo ground truth for detectors on any dataset with single image homography adaptation (HA). Homography warping pairs are generated on-the-fly for descriptor training \cite{detone_superpoint:_2017}. We put a Gaussian filter on the ground truth heatmap to enable prediction with \texttt{Softargmax}, where $\sigma_{fe} = 0.2$.
For descriptor sparse loss, we have $H_c = H/8$, $W_c = W/8$, 
$N = 600$, and $M = 100$. NMS window size is set to be $w = 4$.
% For the PE module, we initialize the PE module based on the correspondences from the freezed FE.
 The model is trained with 200k iterations on synthetic datasets, and 50k iterations on real images.

With the correspondences from the pre-trained FE, we initialize the PE module using \textbf{F-loss}. The network has RNN structure with iteration D=5. The training converges at around 20k iterations. For training with the \textbf{pose-loss}, We set $c_r = 0.1$, $c_t = 0.5$ and $\lambda_{rt} = 0.1$.

When connecting the entire pipeline, the gradients from \textbf{pose-loss} flow back through the Pose Estimation(PE) module to update the prediction of weights, as well as the Feature Extraction(FE) module to update the locations of keypoints. The pipeline and supervision is shown in \figref{fig:pip_supervision}.

\begin{figure*}[]
    \centering
    \includegraphics[trim={1cm 1cm 1cm 1cm}, page=2, width=500pt, height=250pt, keepaspectratio]{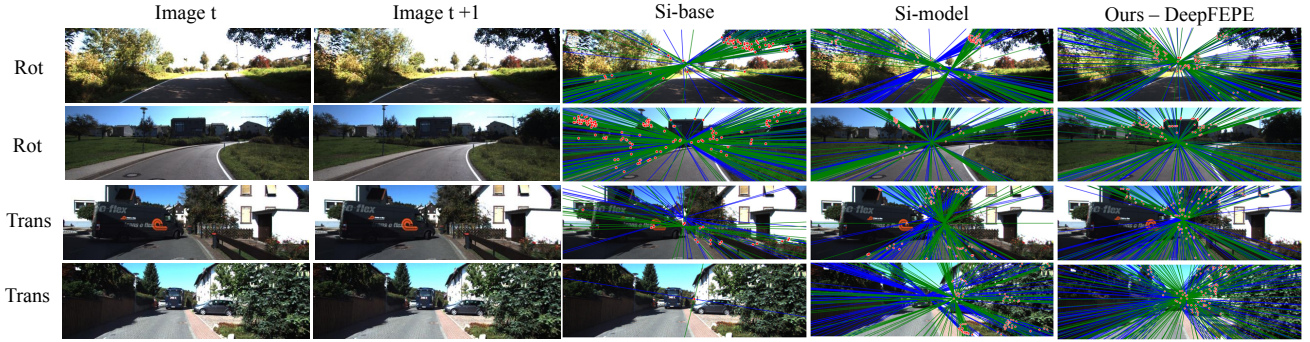}
    % \subfloat[Sequence 09.]{}
    \caption{\textbf{Pose estimation comparison.} The first two columns show the image pairs. The last three columns compare Si-base, Si-model and \texttt{DeepFEPE}. We show 2 examples for rotation and 2 for translation dominated pairs. 
    The blue lines are plotted from ground truth fundamental matrix $F_{gt}$, as the green lines are from estimated $F_{est}$. Red dots are the keypoints from correspondences with high weights.
    For Si-base, the correspondences are selected by RANSAC, where keypoints around vanishing points are usually rejected. However, Si-model utilizes all correspondences to solve for the fundamental matrix, which leads to better quantitative results after training.
    The distribution of points in \texttt{DeepFEPE} is more balanced than for others, leading to more accurate pose estimation.
    % (Lines and dots are plotted with the same way in \figref{fig:vis_poseCp_bad} and \figref{fig:vis_freezeDeepF}.)
    }
    \label{fig:vis_poseCp_good}
\end{figure*}

% ================================ new section =========================================
\section{Experiments} \label{Experiments}
We evaluate \texttt{DeepFEPE} using pose estimation error, and compare with previous approaches. Acronyms and symbols for the approaches are defined in \tabref{tab:exp_ablation_ref_table}. Different methods are evaluated on KITTI dataset \cite{geiger_vision_2013}, and further on ApolloScape dataset\cite{huang_apolloscape_2019} to show the generalization ability to unseen data.
To be noted, we evaluate for relative pose estimation with existing baselines as in \secref{sec:pose_est}, and for visual odometry as in \secref{VO_eval}.
We demonstrate significant improvement quantitatively against baseline learning-based methods after end-to-end training, as shown in \tabref{tab:exp_ablation_kitti-kit-ours} and \tabref{tab:exp_ablation_kitti-apo-ours}. To give an insight into the improvement of optimizing SuperPoint from \textbf{pose-loss}, we evaluate the epipolar error of correspondences quantitatively in \tabref{tab:exp_sp_epiDist} and visualize the change of keypoint distribution during training in \figref{fig:vis_freezeDeepF}. 

\subsection{Datasets}
We extract all pairs of consecutive frames, \ie with time difference 1, for training and testing.\\
\noindent \textbf{KITTI dataset}
We train and evaluate our pipeline using KITTI odometry sequences, with ground truth 6 DoF poses obtained from IMU/GPS. There are 11 sequences in total, where sequences 00-08 are used for training (16k samples) and 09-10 are used for testing (2,710 samples).

\noindent \textbf{ApolloScape dataset}
The dataset is collected in driving scenarios, with ground truth 6 DoF poses collected from GPS/IMU. It includes different view angles of the camera, and lighting variations from KITTI dataset, and is used for generalization testing. We only use testing split in Road 11 for cross-dataset setting (5.8k samples).
% use the training split from Road 12, 14, 15, 16, 17 for training (22k samples), and testing split in Road 11 for testing (5.8k samples).

% The domain gap between KITTI and ApolloScape dataset includes the view angle of camera, the sceneary and lightings in the sequences.

% and evaluate the improvement of Superpoint correspondences using epipolar distance. 

% In order to show our pipeline can generalize and adapt to different scenes, we perform experiments with the following datasets: (1) The KITTI visual odometry dataset, which includes outdoor driving scenes on freeways and in city with static and dynamic objects. The ground truth of camera pose comes from fusion of high-precision IMU and GPS sensors. (2) The TUM RGB-D SLAM dataset, which includes indoor scenes in offices and industrial buildings. The ground truth of camera pose comes from external motion capture system MotionAnalysis.
% (MotionAnalysis, “Raptor-E Digital RealTime System,” http://www.motionanalysis.com/html/industrial/raptore.html.) 

% (3) The community photo collection dataset, which gathered unordered photos from different cameras for a certain landmark. The ground truth is generated from COLMAP.

\subsection{Relative Pose Estimation} \label{sec:pose_est}
We evaluate the performance of \texttt{DeepFEPE} from the estimated rotation and translation, as in \cite{deepf}. With the transformation matrix, we calculate the error by composing the inverse of the ground truth matrix with our estimation. Then, we extract the angle from the composed rotation matrix as the error term.
\begin{eqnarray}
 \mathbf{R}_{rel} &=& \mathbf{R}_{est} * \mathbf{R}_{gt}^T ,  \\    
   \delta \theta &=& \|Rog(\mathbf{R}_{rel})\|_2 , % (rad) 
\end{eqnarray}
where $Rog(.) \in \mathbb{R}^{3\times3} \to \mathbb{R}^{3\times1} $ converts a rotation matrix to a Rodrigues' rotation vector. The length of the resulting vector represents the error in angle. 
We measure translation error by angular error between the estimated translation and the ground truth vector. Due to scale ambiguity, we set the translation vector to be unit vector.
\begin{equation}
   \delta \mathbf{t} = cos^{-1}( \frac{\mathbf{t}_{est} \cdot \mathbf{t}_{gt}}{\| \mathbf{t}_{est}\| \|\mathbf{t}_{gt} \|})  % \tagaddtext{(rad) }
\end{equation}
The equation computes the angle between the estimated translation vector and ground truth vector. 
With the rotation and translation error for each pair of images throughout the sequence, we compute the inlier ratio, from 0\% to 100\%, under different thresholds. Mean and median of the error are also computed in degrees.
% , as well as the mean and median. The unit of inlier ration is percentage, from 0\% to 100\%, the higher the better. The unit of the mean and median is degree, the lower the better. 

We compare different models as follows. 
\textbf{(1)} \textbf{Si-base} (classic baseline models): Correspondences from SIFT are fed into RANSAC for pose estimation.
% : The correspondences are predicted from SIFT\cite{sift} or Superpoint\cite{detone_superpoint:_2017}, and are used in RANSAC to predict poses. 
\textbf{(2)} \textbf{Si-models} (SIFT and DeepF models): SIFT correspondences are used to estimate pose using deep fundamental matrix \cite{deepf}, which is the current state-of-the-art relative pose estimation pipeline.
% : We slightly improve the performance of the model by combining original fundamental matrix loss and our proposed loss. 
\textbf{(3)} \textbf{Sp-base} (SuperPoint with RANSAC): SuperPoint is pre-trained and then connected to RANSAC for pose estimation.
\textbf{(4)} \textbf{Sp-models} (SuperPoint and DeepF models): SuperPoint is pre-trained on the given dataset and then frozen to train DeepF models. 
\textbf{(5)} \texttt{DeepFEPE} (Our method with end-to-end training): feature extraction (FE) and pose estimation (PE) are trained jointly using \textbf{F-loss} or \textbf{pose-loss}. The reference table of the models above are shown in \tabref{tab:exp_ablation_ref_table}, with symbols representing different training combinations. The models are trained on KITTI and evaluated on both KITTI and ApolloScape datasets. 
% The reults are shown in Table \ref{tab:exp_ablation_kittiModels}. The results with models trained on Apollo dataset are shown in Table \ref{tab:exp_ablation_apolloModels}.

%%% show ours on kitti
\tabref{tab:exp_ablation_kitti-kit-ours} compares the learning-based baseline (Sp-base) with our \texttt{DeepFEPE} model, which shows significant improvement \wrt rotation and translation error. Looking into the rotation error, the pre-trained SuperPoint \cite{detone_superpoint:_2017} performs poorly with RANSAC  pose estimation (0.217 degrees median error), whereas the DeepF \cite{deepf} module improves that to 0.078 degrees. Our \texttt{DeepFEPE} further improves the rotation median error to 0.041 degrees, with translation median error from 2.1 (Sp-Ran) to 0.5 degrees.

%%%% kitti model (main table) on kitti

%%%% ours - kitti
% table: table_kittiModels_ours_kit
\begin{table}[ht]
    \scriptsize
    \centering
    \begin{tabular}{lllllll}
        \toprule
        \multirow{2}{*} {KITTI Models} & \multicolumn{6}{c}{KITTI dataset - error(deg.) inlier ratio$\uparrow$, mean$\downarrow$, median$\downarrow$}  \\
        \cmidrule(r){2-7} 
        {}                               & \multicolumn{3}{c}{Rotation (deg.)} & \multicolumn{3}{c}{Translation (deg.)} \\
         {} & 0.1$\uparrow$  & Mean.$\downarrow$ & Med.$\downarrow$ & 2.0$\uparrow$ & Mean.$\downarrow$ & Med.$\downarrow$ \\
        \midrule
Base(Sp-Ran) & 0.189 & 0.641 & 0.217 & 0.481 & 5.798 & 2.103
\\ \hline
Sp-Df-f & 0.633 & 0.100 & 0.078 & 0.830 & 1.476 & 0.846
\\ \hline
Sp-Df-p & 0.875 & 0.130 & 0.047 & 0.887 & 1.719 & 0.539
\\ \hline
Ours(Sp-Df-f-end) & 0.915 & 0.053 & 0.042 & 0.905 & 1.662 & \textbf{0.489}
\\ \hline
Ours(Sp-Df-p-end) & \textbf{0.932} & \textbf{0.050} & \textbf{0.041} & 0.905 & 1.600 & 0.503
\\ \hline
Ours(Sp-Df-fp-end) & 0.910 & 0.054 & 0.048 & \textbf{0.917} & \textbf{1.062} & 0.504
\\ \hline

\bottomrule
    \end{tabular}
\caption{\label{tab:exp_ablation_kitti-kit-ours} \textbf{Comparison of pose estimation for learning-based KITTI models on KITTI dataset.} 
The set of models are trained on KITTI with learning-based feature extraction (FE). 
(Refer to \tabref{tab:exp_ablation_ref_table} for acronyms.)}
% It shows significant improvement from RANSAC to DeepF pose estimation, and from separate models to end-to-end trained models. 
% The set of models are all trained in KITTI with Superpoint as feature extractor. The evaluation of relative pose prediction shows substantial improvement from the Superpoint with deepF baseline(Sp-Df-f) with 30\% for rotation inlier ration and 7\% for translation. The mean rotation and translation error also largely improves due to the end-to-end training.
\end{table}

%%%% baselines - kitti
% table: table_kittiModels_baselines_kit
\begin{table}[h]
    \scriptsize
    \centering
    \begin{tabular}{lllllll}
        \toprule
        \multirow{2}{*} {KITTI Models} & \multicolumn{6}{c}{KITTI dataset - error(deg.) inlier ratio$\uparrow$, mean$\downarrow$, median$\downarrow$}  \\
        \cmidrule(r){2-7} 
        {}                               & \multicolumn{3}{c}{Rotation (deg.)} & \multicolumn{3}{c}{Translation (deg.)} \\
         {} & 0.1$\uparrow$  & Mean.$\downarrow$ & Med.$\downarrow$ & 2.0$\uparrow$ & Mean.$\downarrow$ & Med.$\downarrow$ \\
        \midrule
Base(Si-Ran) & 0.818 & 0.391 & 0.056 & 0.899 & 1.895 & 0.639
\\ \hline
Si-Df-f & 0.938 & \textbf{0.051} & 0.041 & 0.914 & 1.699 & \textbf{0.484}
\\ \hline
Si-Df-p & 0.901 & 0.059 & 0.044 & 0.903 & 1.472 & 0.513
\\ \hline
Si-Df-fp & \textbf{0.947} & 0.111 & \textbf{0.038} & 0.916 & 1.741 & 0.484
\\ \hline
Ours(Sp-Df-fp-end) & 0.910 & 0.054 & 0.048 & \textbf{0.917} & \textbf{1.062} & 0.504
\\ \hline

\bottomrule
    \end{tabular}
\caption{\label{tab:exp_ablation_kitti-kit-base} \textbf{Comparison of pose estimation for SIFT-based KITTI models on KITTI dataset.} The table compares our \texttt{DeepFEPE} model with Si-base and Si-models for pose estimation. 
(Refer to \tabref{tab:exp_ablation_ref_table} for acronyms.) }
% Our model works better than Si-base, and comparable with the state-of-the-art Si-models. 
\end{table}

%%%% kitti models - ours on apollo
% table: table_kittiModels_our_apo
\begin{table}[h]
    \scriptsize
    \centering
    \begin{tabular}{lllllll}
        \toprule
        \multirow{2}{*} {KITTI Models} & \multicolumn{6}{c}{Apollo dataset - error(deg.) inlier ratio$\uparrow$, mean$\downarrow$, median$\downarrow$}  \\
        \cmidrule(r){2-7}
        {}                               & \multicolumn{3}{c}{Rotation (deg.)} & \multicolumn{3}{c}{Translation (deg.)} \\
         {} & 0.1$\uparrow$  & Mean.$\downarrow$ & Med.$\downarrow$ & 2.0$\uparrow$ & Mean.$\downarrow$ & Med.$\downarrow$ \\
        \midrule
Base(Sp-Ran) & 0.407 & 0.205 & 0.118 & 0.583 & 5.645 & 1.670
\\ \hline
Sp-Df-f & 0.725 & 0.126 & 0.068 & 0.754 & 2.074 & 1.155
\\ \hline
Sp-Df-p & 0.730 & 0.124 & 0.067 & 0.827 & 1.905 & 0.974
\\ \hline
Ours(Sp-Df-f-end) & 0.841 & 0.100 & 0.051 & 0.910 & \textbf{1.122} & \textbf{0.589}
\\ \hline
Ours(Sp-Df-p-end) & 0.686 & 0.152 & 0.071 & 0.747 & 2.652 & 1.068
\\ \hline
Ours(Sp-Df-fp-end) & \textbf{0.864} & \textbf{0.092} & \textbf{0.051} & \textbf{0.924} & 1.275 & 0.659
\\ \hline
\bottomrule
    \end{tabular}
\caption{\label{tab:exp_ablation_kitti-apo-ours} \textbf{Comparison of pose estimation for learning-based KITTI models on Apollo dataset.} The table compares the learning-based approaches in a cross-dataset setting. }
% The table shows that our end-to-end \texttt{DeepFEPE} performs the best.
\end{table}

%%%% kitti models - baseline on apollo
% table: table_kittiModels_baselines_apo
\begin{table}[h]
    \scriptsize
    \centering
    \begin{tabular}{lllllll}
        \toprule
        \multirow{2}{*} {KITTI Models} & \multicolumn{6}{c}{Apollo dataset - error(deg.) inlier ratio$\uparrow$, mean$\downarrow$, median$\downarrow$}  \\
        \cmidrule(r){2-7}
        {}                               & \multicolumn{3}{c}{Rotation (deg.)} & \multicolumn{3}{c}{Translation (deg.)} \\
         {} & 0.1$\uparrow$  & Mean.$\downarrow$ & Med.$\downarrow$ & 2.0$\uparrow$ & Mean.$\downarrow$ & Med.$\downarrow$ \\
        \midrule

Base(Si-Ran) & \textbf{0.922} & 0.157 & \textbf{0.037} & \textbf{0.979} & \textbf{0.788} & 0.388
\\ \hline
Si-Df-f & 0.845 & 0.172 & 0.043 & 0.895 & 2.452 & 0.389
\\ \hline
Si-Df-p & 0.727 & 0.333 & 0.056 & 0.760 & 4.918 & 0.658
\\ \hline
Si-Df-fp & 0.840 & 0.148 & 0.044 & 0.911 & 2.103 & \textbf{0.369}
\\ \hline
Ours(Sp-Df-fp-end) & 0.864 & \textbf{0.092} & 0.051 & 0.924 & 1.275 & 0.659
\\ \hline

\bottomrule
    \end{tabular}
\caption{\label{tab:exp_ablation_kitti-apo-base} \textbf{Comparison of pose estimation for SIFT-based KITTI models on Apollo dataset.} The table compares our \texttt{DeepFEPE} with other baseline methods in a cross-dataset setting. }
% The results show that Si-base maintains the best overall results, and \texttt{DeepFEPE} performs better than Si-models.
\end{table}

\begin{figure}[]
    \centering
    \includegraphics[trim={0.5cm 0cm 0.5cm 0cm}, page=2, width=240pt, height=150pt, keepaspectratio]{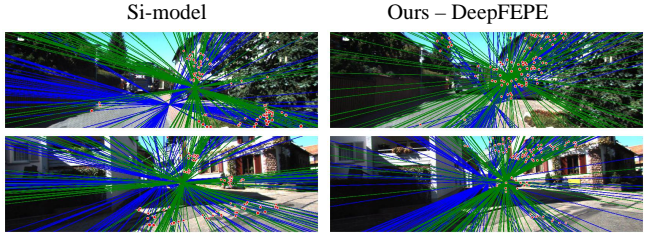}\label{fig:fig_point_scores}
    % \subfloat[Sequence 09.]{}
    \caption{\textbf{Failure cases of pose estimation.} The failure cases include over-exposed and textureless scenes. The challenging views result in noisy correspondences and wrong predictions. (Lines and dots are plotted as in \figref{fig:vis_poseCp_good}.)
    }
    % Compare Si-model with our \texttt{DeepFEPE}.
    %  scenes which include more rotation motion are harder than forward translation cases.
    \label{fig:vis_poseCp_bad}
\end{figure}

In terms of other baselines, we compare \texttt{DeepFEPE} with Si-models and Si-base in \tabref{tab:exp_ablation_kitti-kit-base}. \texttt{DeepFEPE} achieves better mean translation and rotation error compared to Si-base, and comparable performance with Si-models. The table demonstrates that the \texttt{DeepFEPE} model sets up the new state-of-the-art for learning-based relative pose estimation against DeepF. The qualitative results are shown in \figref{fig:vis_poseCp_good} and \figref{fig:vis_poseCp_bad}, comparing Si-base, Si-model and \texttt{DeepFEPE}. Pose estimation is visualized by comparing the epipolar lines projected from ground truth and estimated fundamental matrices. If the estimated one is close to ground truth, the vanishing point should match that of ground truth. Keypoints with high weights predicted by Pose Estimation (PE) are also plotted for reasoning the relation of point distribution and pose estimation.

%%% show cross dataset
Due to the fact that learning-based methods are biased towards the training data, we evaluate the models trained from KITTI on the ApolloScape dataset. The results demonstrate that our model retains generalization ability and is less prone to overfitting.
From \tabref{tab:exp_ablation_kitti-apo-ours}, we compare \texttt{DeepFEPE} models with Sp-base models and observe the benefit from end-to-end training with lower rotation and translation error. Without end-to-end training, the Sp-base models degrade significantly (in \tabref{tab:exp_ablation_kitti-kit-ours}) and are won over by end-to-end models by a large margin. Compared to other baselines in \tabref{tab:exp_ablation_kitti-apo-base}, we observe that the Si-base demonstrates the highest accuracy, and \texttt{DeepFEPE} achieves better mean rotation and translation error over Si-models.

\begin{figure}[]
    \centering
    \includegraphics[trim={1cm 1cm 1cm 1cm}, page=2, width=240pt, height=200pt, keepaspectratio]{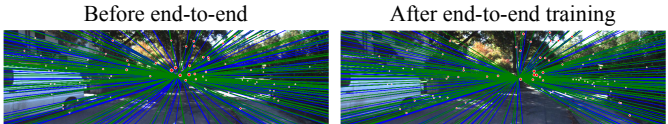} % \label{fig:fig_point_scores}
    % \subfloat[Sequence 09.]{}
    \caption{\textbf{Change of keypoint distribution after end-to-end training.} To show the qualitative results of feature extractor, we freeze the pose estimation module and update only the feature extractor. (Lines and dots are plotted as in \figref{fig:vis_poseCp_good}.)}
    % In our pipeline, we train feature extractor and pose estimation end-to-end. 
    % The results show that the change of distribution helps the pose estimation predict better pose. 
    \label{fig:vis_freezeDeepF}
\end{figure}

%%% show the benefit of pose loss
To further examine the benefit of geometry-based loss, we can look into \tabref{tab:exp_ablation_kitti-kit-ours}, with 3 models trained on either \textbf{F-loss}, \textbf{pose-loss} or both. We can observe the model trained using both losses achieves significantly better mean translation error. We believe this is because the geometric information incorporated in \textbf{pose-loss} encourages the keypoint distribution in FE to be pose-aware. The keypoints are updated to balance between good localization accuracy and matching \wrt the pose estimation. The change of keypoint distribution is observed from \figref{fig:vis_freezeDeepF}. This shows the potential of having a robust and optimized feature extractor with end-to-end training. As observed from the figure, keypoints close to the vanishing point are reduced after the end-to-end training. It is because these points are good for matching but may incur high triangulation errors when solving for camera pose, due to their little motion from frame to frame. On the other hand, points near the border of the image see a noticeable increase. These points may not be robustly matched with conventional descriptors because of large motion and in some cases motion blur. On the contrary, our method is able to reveal these points which provide a wider baseline for more accurate camera pose estimation.

%%%%%%%%%%%%%%%%%%%%%%%%%%%%%%%%%%%%%%%%%   end  %%%%%%%%%%%%%%%%%%%%%%%%%%%%%%%%%%%%%%%%%%%%%%%%%%%%%%%%

%%%%%%%%%%%%%%%%%%%%%%%%%%%%%%%%%%%%%%%%% start sp training %%%%%%%%%%%%%%%%%%%%%%%%%%%%%%%%%%%%%%%%%%%%%
% \begin{figure}[!tbp]
%   \centering
%   \subfloat[Sequence 09.]{\includegraphics[page=1, width=110pt, height=110pt]{latex/imgs/vis_freezeDeepF.png}\label{fig:f1}}
%   \hfill
%   \subfloat[Sequence 10.]{\includegraphics[page=1, width=110pt, height=110pt]{latex/imgs/vis_freezeDeepF.png}\label{fig:f2}}
%   \caption{KITTI trajectories of pose estimation results. No ground truth or scale fixing are used in testing sequences.}
%   \label{fig:traj_kitti}
% \end{figure}

%%% corrs
% \begin{figure}[]
%     \centering
%     % \includegraphics[width=8cm]{latex/imgs/pipeline1.png}
%     \includegraphics[width=240pt, height=200pt, keepaspectratio]{imgs/vis_corr_rand_0-3_freezeDeepF_highlighted.png} % \label{fig:fig_point_scores}
%     % \subfloat[Sequence 09.]{}
%     \caption{\textbf{Change of correspondences during training.} We visualize the correpondence distribution along training (0, 400, 1k iterations). We plot out the dots and correpondences. All the green dots are detected matching, and we only show 30\% of the correpondences due to the large volume. We labeled the correpondences added as big yellow circles, and correspondences removed as big red circles.}
%     \label{fig:vis_freezeDeepF_corrs}
% \end{figure}
%%%%%%%%%%%%%%%%%%%%%%%%%%%%%%%%%%%%%%%%%   end  %%%%%%%%%%%%%%%%%%%%%%%%%%%%%%%%%%%%%%%%%%%%%%%%%%%%%%%%

\subsection{Visual Odometry Evaluation} \label{VO_eval}
%%% show kitti VO
To evaluate the prediction over the whole sequence, we test the models on KITTI sequences 09, 10. We set the relative translation vectors to be unit vectors and align the trajectory with the ground truth using Sim(3) Umeyama alignment\cite{kitti, NIPS2019_scsfmlearner}\footnote{https://github.com/Huangying-Zhan/kitti-odom-eval}. We compute the relative translation and rotation error, as shown in \tabref{tab:exp_kitti_metric}. Our \texttt{DeepFEPE} model significantly improves the SuperPoint baseline and works comparable with the Si-base pipeline.

%%%% ours - kitti
% table: table_kittiModels_ours_kit
\begin{table}[ht]
    \scriptsize
    \centering
    \begin{tabular}{lllll}
        \toprule
        % \multirow{2}{*} {Methods} & \multicolumn{4}{c}{KITTI dataset - error(deg.)}\\
        % \cmidrule(r){2-5} 
        \multirow{2}{*} {Methods}     & \multicolumn{2}{c}{Seq. 09} & \multicolumn{2}{c}{Seq. 10} \\
        {} & $t_{err}(\%)$ & $r_{err}(\deg / 100m)$ & $t_{err}$ & $r_{err}$ \\
    %     %  {} & 0.1$\uparrow$  & Mean.$\downarrow$ & Med.$\downarrow$ & 2.0$\uparrow$ & Mean.$\downarrow$ & Med.$\downarrow$ \\
        \midrule
Base(Si-Ran) & 8.842  & 0.512 & 11.508 & 1.447 \\ \hline
Base(Sp-Ran) & 11.005 & 3.324 & 40.021 & 29.599 \\ \hline
Si-Df-fp    & 9.706 & 0.889 & 11.206 & 1.546 \\ \hline
Ours(Sp-Df-fp-end)    & 8.639 & 0.664 & 11.719 & 0.945

\\ \hline

\bottomrule
    \end{tabular}
\caption{\label{tab:exp_kitti_metric} \textbf{Visual odometry results on KITTI.} }
% The table compare our \texttt{DeepFEPE} model with Si-base and Si-models for pose estimation.  
\end{table}

\begin{table}[h]
    \scriptsize
    \centering
    \begin{tabular}{lllllllllllllll}
        \toprule
        \multirow{2}{*} {} & \multicolumn{7}{c}{KITTI - epiploar dists (n pixels), num of matches: mean/ med.}  \\
        \cmidrule(r){2-8} \cmidrule(r){9-15} \\
        KITTI Models  & 0.2 & 0.5 & 1.0 & 2.0  & Mean & Med. \\
        \midrule
        
Sp-Ran & 0.080 & 0.195 & 0.361 & 0.581 & 541.546 & 533.000
\\ \hline
Sp-Df-f-end & 0.107 & 0.258 & 0.460 & 0.685 & 719.986 & 703.000
\\ \hline
Sp-Df-p-end & 0.096 & 0.232 & 0.421 & 0.643 & 626.343 & 611.000
\\ \hline
Sp-Df-fp-end & 0.105 & 0.254 & 0.453 & 0.677 & 669.170 & 654.000
\\ \hline

        % Apollo Models & 0.2 & 0.5 & 1.0 & 2.0 & 5.0 & Mean(2.0) & Med.(2.0) & 0.2 & 0.5 & 1.0 & 2.0 & 5.0 & Mean(2.0) & Med.(2.0) \\
\bottomrule
    \end{tabular}
\caption{\label{tab:exp_sp_epiDist} \textbf{Superpoint evaluation.} }
% When training end-to-end, we can see the increase of inlier ratio w.r.t. Sampson distance and number of correspondences. 
\end{table}

\subsection{SuperPoint Correspondence Estimation}

To understand how the Feature Extraction (FE) module is updated after training, we collect quantitative results using Sampson distance and demonstrate keypoint distribution qualitatively.
For each pair of correspondences, we calculate the Sampson distance from \equref{eq:sampson}, which indicates whether the pair of points lies close to each other's epipolar line. We show the inlier ratio \wrt different distance value (unit: pixel) from 0.2 to 2, as well as the number of correspondences in \tabref{tab:exp_sp_epiDist}. The results show an increase of inlier ratio up to 10\% with Sampson distance below 1px on KITTI dataset. The number of correspondences also increases by around 20\%. The result shows that the end-to-end training improves the individual module as well. The model trained on \textbf{F-loss} has the best result under this metric, as the \textbf{F-loss} minimizes the energy in the epipolar space.

% From table \ref{tab:exp_sp_epiDist}, 
% we evaluate correspondences from superpoint model using epipolar distance. Given a ground truth pose and calibration matrix, we can get fundamental matrix. We calculate the inlier ratio of the correspondences that lies in each other's epipolar line within certain distance. 
% We compare the models before and after end-to-end training from pose loss. \textbf{The results show an increase of inlier ratio (around 5\% with epipolar distance 1 px) on KITTI dataset. The number of all correspondences, inlier correspondences increases substantially, and histograms are shown in supplementary material.} The results show the improvement of training end-to-end using pose loss.

% To better understand the keypoint distribution of SuperPoint in the pipeline along training, we freeze DeepF model and train solely on SuperPoint with pose loss. We show the change of keypoint distribution, by plotting the weights of keypoints with low Sampson distance, in \figref{fig:vis_freezeDeepF_corrs}. Along the training, the number of correspondences increases as well as some points are removed or updated.
% keypoints spread out from the region very close to the vanishing point, toward balanced distribution. The example shows that the balanced distribution results in more accurate pose estimation.

% \input{latex/tables/exp_ablation_kitti_all.tex}
% \input{tables/exp_ablation_kittiModels.tex}

% \input{tables/exp_ablation_apolloModels.tex}

% \import{sections/}{ablation.tex}
\section{Conclusion} \label{Conclusion}

In this paper we propose an end-to-end trainable pipeline for estimating camera poses from input image pairs. We demonstrate that our performance is on par with classic methods, and superior generalization ability to unseen data compared with existing baselines. Both qualitative and quantitative results are included in the paper to support the claim. We provide further insights into the benefits that end-to-end training brings into keypoint detection, feature extraction and pose estimation. Future work of this paper may include sequential input or keyframes with long-term temporal cues. Experiments on other datasets with different motion patterns than the driving datasets in the paper can also be explored.

\section*{ACKNOWLEDGMENT}
We thank Ishit Mehta for helpful discussions. This work is supported by NSF CAREER Award 1751365.

% ================================ new section =========================================
% \section{Discussion} \label{Discussion}
% State that we are doing correctly and show it makes sense.
% Deep fundamental matrix with pose loss.
% Superpoint + Deep fundamental matrix with pose loss.

% % ================================ new section =========================================
% \section{Ablation Study} \label{Ablation}
% Deep fundamental matrix with pose loss.
% Superpoint + Deep fundamental matrix with pose loss.

% \input{latex/tables/supp_ablation_kittiModels.tex}
% \input{latex/tables/supp_ablation_apolloModels.tex}
% \input{latex/tables/supp_sp_epiDist.tex}
% ================================ new section =========================================

%-------------------------------------------------------------------------
%-------------------------------------------------------------------------
%----------------------- end of paper ------------------------------------
%-------------------------------------------------------------------------
%-------------------------------------------------------------------------

{\small
\bibliographystyle{IEEEtran}
% \bibliography{DeepSfm_cvpr20, other_references}
\bibliography{egbib}
}

\end{document}